\documentclass[10pt,twocolumn,letterpaper]{article}
\usepackage[accsupp]{axessibility}
\usepackage{iccv}
\usepackage{times}
\usepackage{epsfig}
\usepackage{graphicx}
\usepackage{amsmath,nccmath}
\usepackage{amssymb}
\usepackage{amsmath,amsthm,mathtools}
\usepackage{multirow} 
\usepackage{microtype}
\usepackage{graphicx}
\usepackage{booktabs}
\usepackage{bm}
\usepackage{enumitem}

\usepackage{bbold}
\usepackage[table]{xcolor}
\usepackage{xcolor}
\usepackage{multirow}
\usepackage{subfig}
\usepackage{rotating}
\usepackage{kantlipsum}
\usepackage[symbol]{footmisc}
\usepackage{comment}
\usepackage{xcolor}
\usepackage[pagebackref=true,breaklinks=true,letterpaper=true,colorlinks,bookmarks=false]{hyperref}

 \iccvfinalcopy 

\def\iccvPaperID{9100} 

\newcommand{\figintro}{
\begin{figure}[t]
\setlength\belowcaptionskip{-0.5\baselineskip}
    \centering
    \includegraphics[width=1\columnwidth]{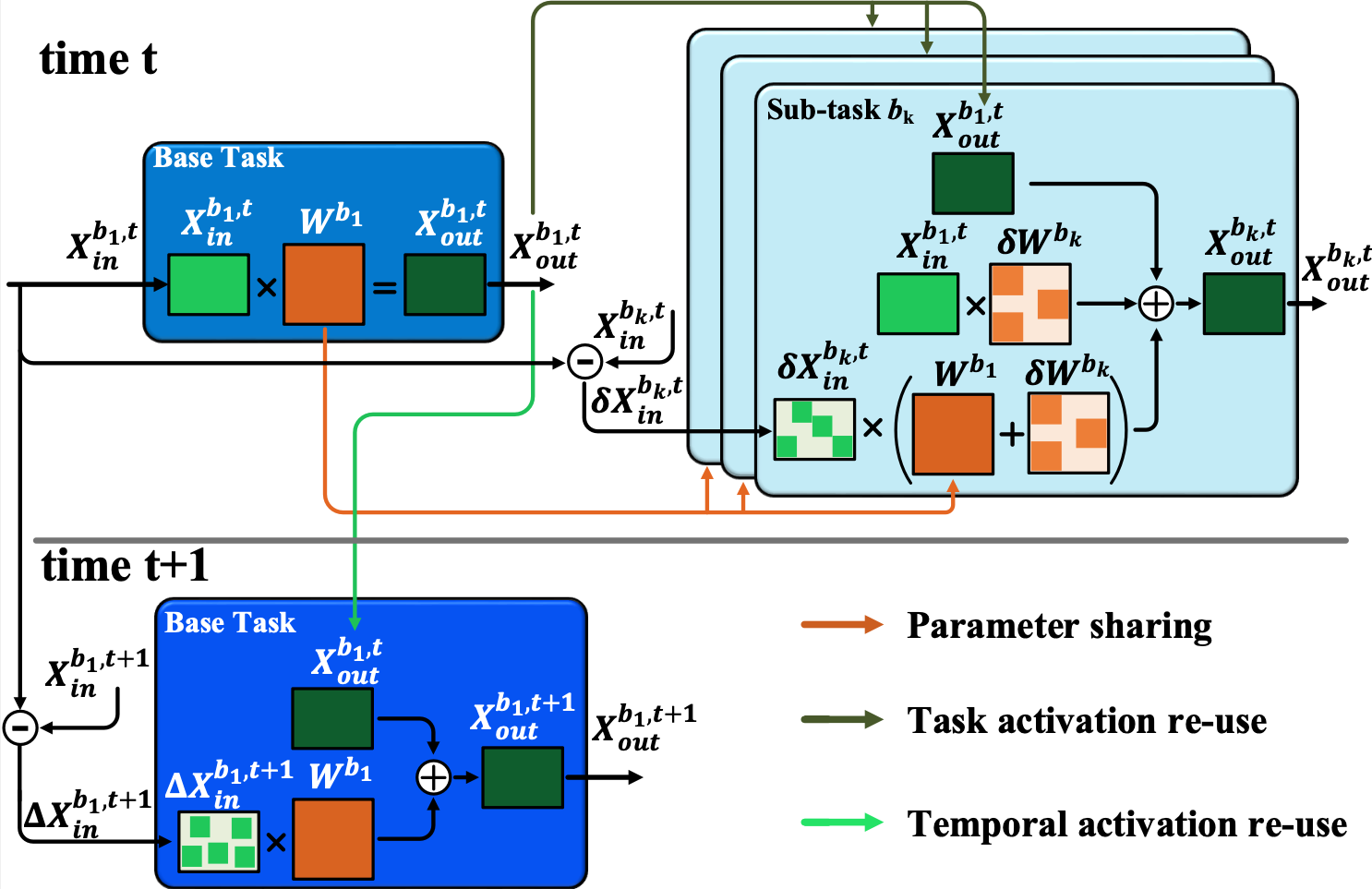}
    \caption{Parameter and activation sharing across task and temporal domains. First, the base task produces its activations at time $t$, then passes them to sub-tasks $b_k$ to share the computation across the task domain. Also, the base task activations at time $t$ are shared with time $t+1$ to  reduce the computation across the temporal domain.}
    \label{fig:fig_intro}
\end{figure}
}

\newcommand{\figvis}{
\begin{figure*}[t]
    \centering
    \includegraphics[width=0.9\linewidth]{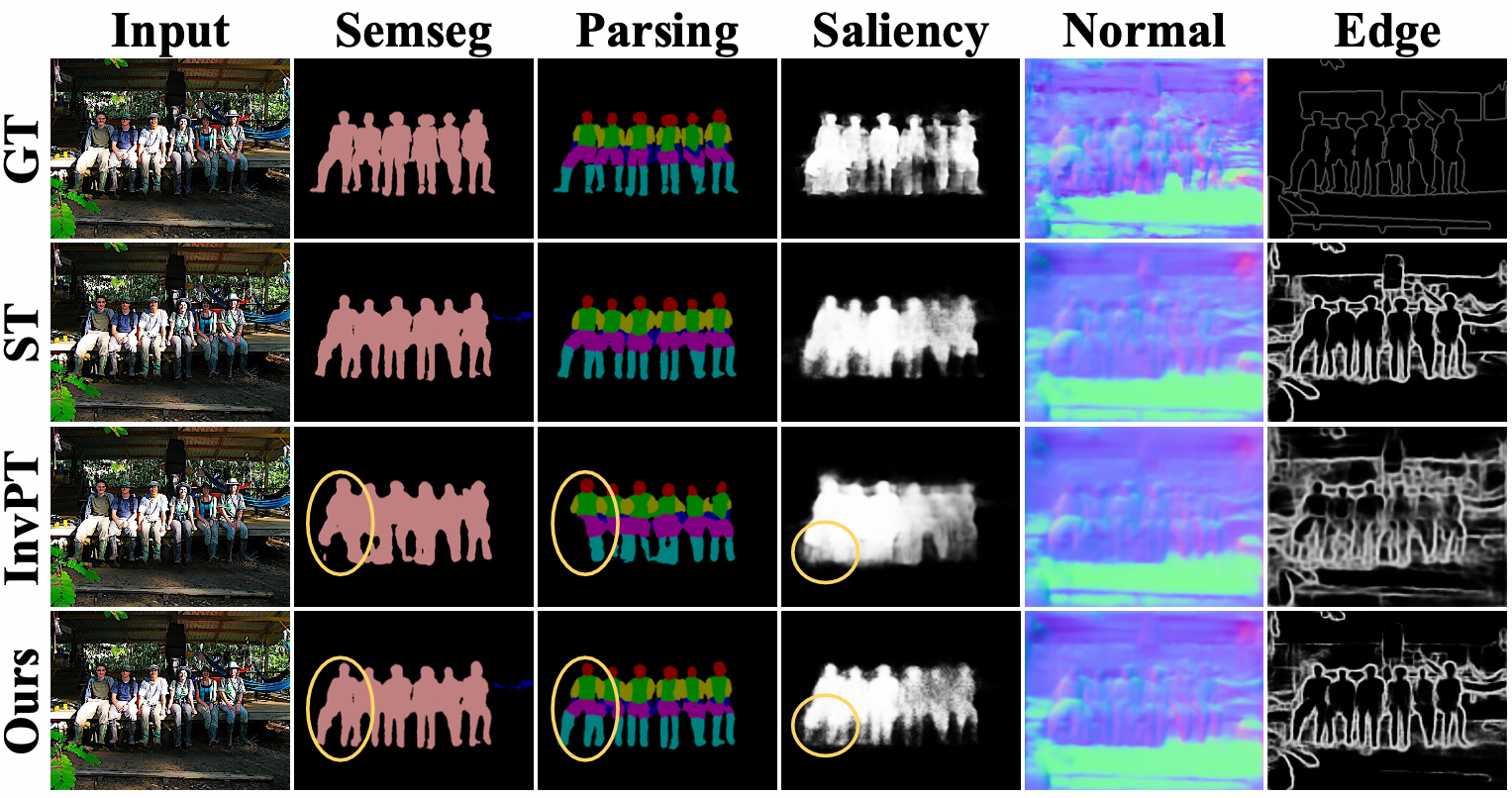}
    \caption{Qualitative comparison with the previous SOTA InvPT and `ST' model on PASCAL-Context dataset. Examples of the regions where our model outperforms InvPT are shown with yellow circles. }
    \label{fig:fig_pas_vis}
\end{figure*}
}

\newcommand{\figwholepasfinal}{
\begin{figure*}[t]
\setlength\belowcaptionskip{-1\baselineskip}
    \centering
    \begin{minipage}{.47\textwidth}
        \centering
        \includegraphics[width=1\columnwidth]{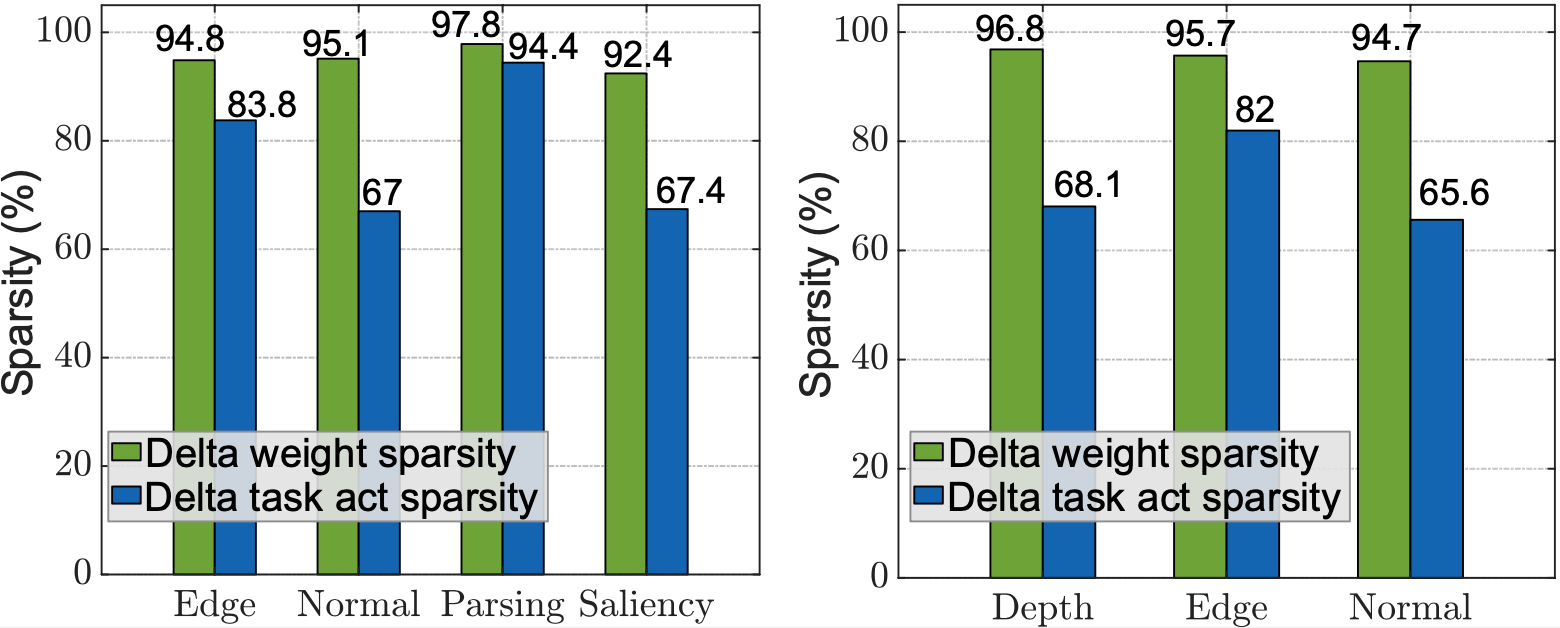}
    \caption{Overall delta weight and delta task activation sparsity for the Pascal-Context dataset ($\emph{left}$) and the NYUD-v2 dataset ($\emph{right}$).}
    \label{fig:all_sparsity}
    \end{minipage} \qquad
    \begin{minipage}{0.48\textwidth}
       \centering
        \includegraphics[width=1\columnwidth]{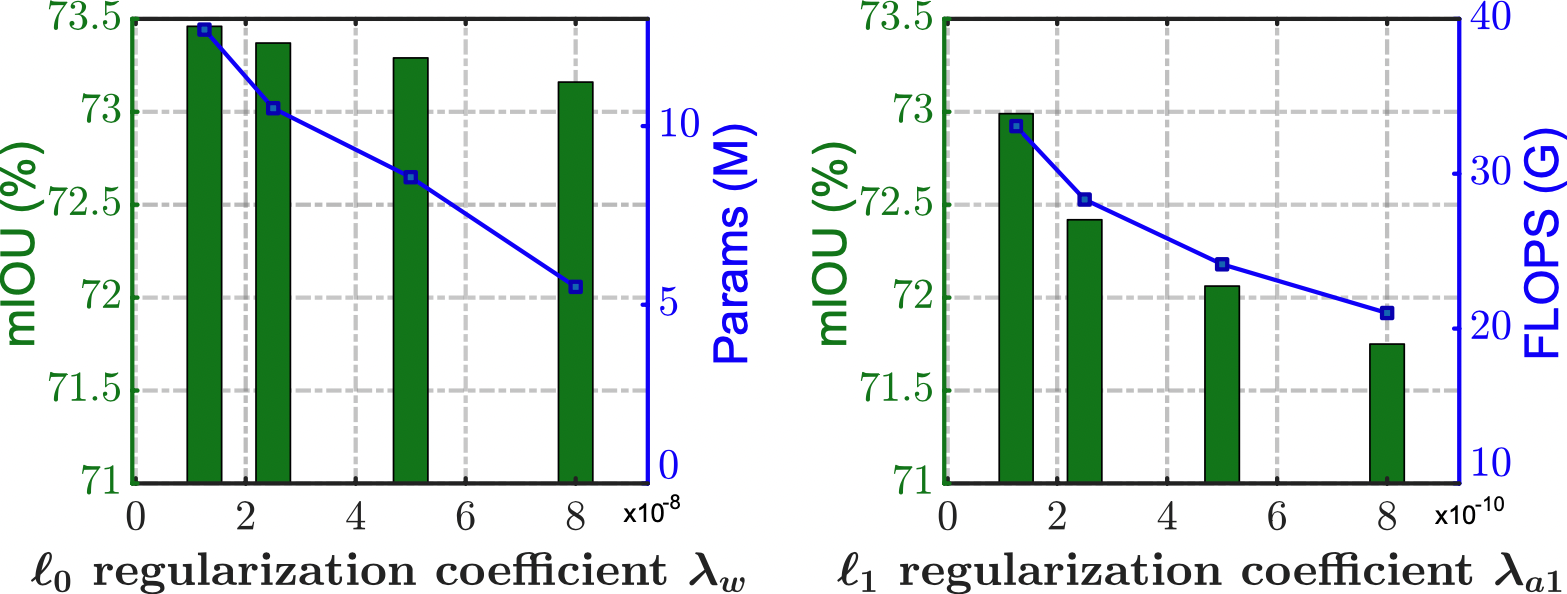}
        \caption{The impacts of $\lambda_w$ and $\lambda_{a1}$ on the human parsing task for Pascal-context dataset.}
    \label{fig:fig_abl_task}
    \end{minipage}
\end{figure*}
}

\newcommand{\tabelablationnpascal}{
    \begin{table*}[t]
    \caption{Performance and complexity of our proposed method with different levels of sharing on (a) PASCAL-Context dataset and (b) NYUD-v2 for a single image.} 
    \vspace{-4mm}
    \label{table:share_pasca}
    \centering
    \small
    \resizebox{2.1\columnwidth}{!}{%
    \subfloat[Performance results for Pascal-Context dataset.]{%
    \begin{tabular}{c|ccccc|ccccc}
    \hline
    \multirow{2}{*}{\textbf{Method}}        & \textbf{Semseg} & \textbf{Parsing} & \textbf{Saliency} & \textbf{Normal} & \textbf{Boundary} & \multirow{2}{*}{\textbf{Param} (M)} & \multirow{2}{*}{\textbf{Memory} (Mb)}  & \textbf{Total} & \textbf{Base-task} & \textbf{Per Sub-task}                \\
           & \scriptsize \textbf{mIoU} $\uparrow$  & \scriptsize \textbf{mIoU} $\uparrow$  & \scriptsize \textbf{maxF} $\uparrow$    & \scriptsize \textbf{mErr} $\downarrow$     & \scriptsize \textbf{odsF} $\uparrow$ &  & & \textbf{FLOPs} (G) & \textbf{FLOPs} (G) & \textbf{FLOPs} (G)\\ \hline\hline
           \\[\dimexpr-\normalbaselineskip+2pt]
     Single Task (ST) & 78.24 &	73.85&	85.33&	13.32&	79.9&	438.02&	1776 & 585.15 & 120.53 & 116.15      \\ \hline
    \\[\dimexpr-\normalbaselineskip+2pt]
    InvPT  & 77.33  & 66.62 & 85.14 & 13.78 & 73.2 & 176.35 & 705 & 412.17 & 82.43 & 82.43  \\
    Ours + weight sharing        & 78.24 & 73.16 & 85.42 & 13.34 & 79.9 & 114.02 & 482 & 585.15  & 120.53 & 116.15 \\
    \rowcolor[gray]{.90}
        Ours + weight + act sharing   & \textbf{78.24}  & \textbf{71.71} & \textbf{85.28}  & \textbf{13.65} & \textbf{78.6} & \textbf{114.02} & \textbf{482} & \textbf{295.84}& 120.53 & \textbf{43.76}\\
     \hline
    \end{tabular}}}
    \resizebox{1.95\columnwidth}{!}{%
    \subfloat[Performance results for NYUD-v2 dataset.]{%
    \begin{tabular}{c|cccc|ccccc}
    \hline
    \multirow{2}{*}{\textbf{Method}}        & \textbf{Semseg} & \textbf{Depth} & \textbf{Normal} & \textbf{Boundary} & \multirow{2}{*}{\textbf{Param} (M)} & \multirow{2}{*}{\textbf{Memory} (Mb)}   & \textbf{Total} & \textbf{Base-task} & \textbf{Per Sub-task}                    \\
           & \scriptsize \textbf{mIoU} $\uparrow$  &  \scriptsize \textbf{RMSE} $\downarrow$    & \scriptsize \textbf{mErr} $\downarrow$     & \scriptsize \textbf{odsF} $\uparrow$ &  &  & \textbf{FLOPs} (G) & \textbf{FLOPs} (G) & \textbf{FLOPs} (G) \\ \hline\hline
           \\[\dimexpr-\normalbaselineskip+2pt]
    Single Task (ST) & 50.48 &		521.5& 17.26&	77.84&		366.64 & 1452&	314.14  & 75.20 & 79.64        \\ \hline
    \\[\dimexpr-\normalbaselineskip+2pt]
    InvPT  &  50.30  & 536.7 & 19.00 & 77.6 &  160.57 & 643  & 229.43 & 57.35 & 57.35  \\
    Ours + weight sharing        & 50.46 &  529.6& 17.62 & 77.84 &  122.46 & 477 & 314.14  & 75.20 & 79.64 \\
    \rowcolor[gray]{.90}
    Ours + weight + act sharing   & \textbf{50.46}  & \textbf{533.2} & \textbf{18.42}  & \textbf{77.89} &  \textbf{122.46} &  \textbf{477} & \textbf{186.88} & 75.20 & \textbf{37.22}   \\
     \hline
    \end{tabular}}}
    \vspace{-7mm}
    \end{table*}
}

\newcommand{\tablesota}{
    \begin{table*}[t]
\caption{Performance comparison on PASCAL-Context (\emph{left}) and on NYUD-v2 (\emph{right}). `$\uparrow$': higher better, `$\downarrow$': lower better.} 
\label{PAS_SOTA}
  \begin{minipage}{1.1\columnwidth}
    
    \centering
    \hrule
    
    \resizebox{\columnwidth}{!}{%
      \begin{tabular}{c|ccccc}
    \hline
    \multirow{2}{*}{\textbf{Method} }       & \textbf{Semseg} & \textbf{Parsing} & \textbf{Saliency} & \textbf{Normal}& \textbf{Boundary}                     \\
           & \scriptsize \textbf{mIoU} $\uparrow$  & \scriptsize \textbf{mIoU} $\uparrow$  & \scriptsize \textbf{maxF} $\uparrow$    & \scriptsize \textbf{mErr} $\downarrow$     & \scriptsize \textbf{odsF} $\uparrow$    \\ \hline\hline
           \\[\dimexpr-\normalbaselineskip+2pt]
    ASTMT \cite{maninis2019attentive}& 68.00  &  61.10 & 65.70 & 14.70 & 72.40\\
    PAD-Net \cite{xu2018pad}&  53.60 &	 59.60&	65.80&	15.30&	 72.50\\
    MTI-Net \cite{vandenhende2020mti}&  61.70 & 60.18 & 84.78 & 14.23 &  70.80\\
    ATRC \cite{bruggemann2021exploring}&  62.69  &  59.42 &  84.70  & 14.20 & 70.96\\
    ATRC-ASPP \cite{bruggemann2021exploring}& 63.60  & 60.23 & 83.91  & 14.30 & 70.86\\
    ATRC-BMTAS \cite{bruggemann2021exploring}&  67.67  &  62.93 & 82.29  & 14.24 & 72.42\\
    InvPT (ViT-B) \cite{ye2022inverted}&  77.33  &  66.62 & 85.14  & 13.78 & 73.20    \\
    \rowcolor[gray]{.90}
    Ours (ViT-B)   & \textbf{78.24}  & \textbf{71.71} & \textbf{85.28}  & \textbf{13.65} & \textbf{78.6}\\
     \hline
    \end{tabular}
    }
    
  \end{minipage}\hfill 
  \begin{minipage}{0.9\columnwidth}
    \label{NYU_SOTA}
    \centering
    \hrule
    \resizebox{\columnwidth}{!}{%
      \begin{tabular}{c|cccc}
    \hline
    \multirow{2}{*}{\textbf{Method}}        & \textbf{Semseg}& \textbf{Depth} & \textbf{Normal} & \textbf{Boundary}                     \\
           & \scriptsize \textbf{mIoU} $\uparrow$  &   \scriptsize \textbf{RMSE} $\downarrow$    & \scriptsize \textbf{mErr} $\downarrow$     & \scriptsize \textbf{odsF} $\uparrow$    \\ \hline\hline
           \\[\dimexpr-\normalbaselineskip+2pt]
    Cross-Stitch \cite{misra2016cross}& 36.34  &  629.0 & 20.88 & 76.38    \\
    PAP \cite{zhang2019pattern}& 36.72 & 617.8 &	 20.82&	 76.42       \\
    PSD \cite{zhou2020pattern}&  36.69 & 624.6 & 20.87 &  76.42     \\
    PAD-Net \cite{xu2018pad}& 36.61  & 627.0 & 20.85  &  76.38     \\
    MTI-Net \cite{vandenhende2020mti}& 45.97  & 536.5 & 20.27  &  77.86       \\
    ATRC \cite{bruggemann2021exploring}& 46.33  & 536.3 & 20.18  & 77.94       \\
    InvPT (ViT-B) \cite{ye2022inverted}&  50.30  & 536.7 & 19.00  & 77.60    \\
    \rowcolor[gray]{.90}
        Ours (ViT-B)   & \textbf{50.46}  & \textbf{533.2} & \textbf{18.42}  & \textbf{77.89}      \\
     \hline
    \end{tabular}
    }

  \end{minipage}
\end{table*}
}
\newcommand{\tabletmpSOTAwholeNEW}{
    \begin{table*}[t]

  \begin{minipage}{1.05\columnwidth}
  
    \caption{Performance of our method when the base task is segmentation
`Semgseg' or `Parsing` on Pascal-context dataset using Swin-B as the transformer backbone (replacing ViT-B).}
    \centering
    \hrule
    \label{SWINB}
    \resizebox{\columnwidth}{!}{%
      \begin{tabular}{c|ccccc|cc}
    \hline
    \multirow{2}{*}{\textbf{Method}}       & \textbf{Semseg} & \textbf{Parsing} & \textbf{Saliency} & \textbf{Normal}& \textbf{Boundary} & \textbf{Param} &\textbf{FLOPs} \\                   
           & \scriptsize \textbf{mIoU} $\uparrow$  & \scriptsize \textbf{mIoU} $\uparrow$  & \scriptsize \textbf{maxF} $\uparrow$    & \scriptsize \textbf{mErr} $\downarrow$     & \scriptsize \textbf{odsF} $\uparrow$ & \scriptsize(M) &  \scriptsize (G) \\ \hline\hline \\[\dimexpr-\normalbaselineskip+2pt]
    ST & 79.1 & 71.6 & 84.6 & 13.5 & 76.0 & 516.1&444.3\\
    \hline \\[\dimexpr-\normalbaselineskip+2pt]
    InvPT  & 77.5  & 66.8 & 83.6 & 14.6 & 73.0  & 187.2 & 384.7 \\
    \rowcolor[gray]{.90}
    \shortstack{Ours w/Semseg base}        & \textbf{79.1} & 70.0 & \textbf{84.2} & \textbf{14.3} &   \textbf{74.5} & \textbf{149.3} &\textbf{241.1} \\
    \shortstack{Ours w/Parsing base}        & 77.5 & \textbf{71.6} & 84.1 & 14.4 &   74.4 & 149.5 &257.0 \\
    
     \hline
    \end{tabular}
    }
  \end{minipage}\quad
  \vspace{-1mm}
  \begin{minipage}{1\columnwidth}
  
    \caption{Performance of our method using temporal and task activation re-use technique for NYUD-v2 dataset.}
    \centering
    \hrule
    \label{tempPAS_SOTA}
    \resizebox{\columnwidth}{!}{%
      \begin{tabular}{c|cccc|cc}
    \hline
    \multirow{2}{*}{\textbf{Method}}        & \textbf{Semseg} & \textbf{Depth} & \textbf{Normal} & \textbf{Boundary} &  \textbf{Param}&\textbf{FLOPs}                   \\
           & \scriptsize \textbf{mIoU} $\uparrow$  &  \scriptsize \textbf{RMSE} $\downarrow$    & \scriptsize \textbf{mErr} $\downarrow$     & \scriptsize \textbf{odsF} $\uparrow$ &\scriptsize(M) &  \scriptsize(G)  \\ \hline\hline \\[\dimexpr-\normalbaselineskip+2pt]
    ST & 50.48 & 521.5 & 17.26 & 77.84 & 366.64  &			314.14       \\ \hline
    \\[\dimexpr-\normalbaselineskip+2pt]
    InvPT  & 50.30  & 536.7 & 19.00 & 77.6 &160.57 &229.43  \\
    \shortstack{Ours + Task act}        & 50.46  & 533.2 & 18.42 & 77.89 &   122.46 &186.88   \\[\dimexpr-\normalbaselineskip+12pt]
    \rowcolor[gray]{.90}
    \shortstack{Ours   + Task act  \\+ Temporal act}        & \textbf{50.40} & \textbf{532.41} & \textbf{18.42} & \textbf{77.79} &  \textbf{122.46} & \textbf{107.87 } \\
    
     \hline
    \end{tabular}
    }
    
  \end{minipage}
\end{table*}
}

\ificcvfinal\fi

\begin{document}

\title{Efficient Computation Sharing for Multi-Task Visual Scene Understanding}

\author{Sara Shoouri
\qquad
Mingyu Yang
\qquad
Zichen Fan
\qquad
Hun-Seok Kim \\[2mm] University of Michigan \\ {\tt\small \{sshoouri,mingyuy,zcfan,hunseok\}@umich.edu}
}

\maketitle
\ificcvfinal\thispagestyle{empty}\fi

\begin{abstract}
Solving multiple visual tasks using individual models can be resource-intensive, while multi-task learning can conserve resources by sharing knowledge across different tasks. Despite the benefits of multi-task learning, such techniques can struggle with balancing the loss for each task, leading to potential performance degradation. We present a novel computation- and parameter-sharing framework that balances efficiency and accuracy to perform multiple visual tasks utilizing individually-trained single-task transformers. Our method is motivated by transfer learning schemes to reduce computational and parameter storage costs while maintaining the desired performance. Our approach involves splitting the tasks into a base task and the other sub-tasks, and sharing a significant portion of activations and parameters/weights between the base and sub-tasks to decrease inter-task redundancies and enhance knowledge sharing. The evaluation conducted on NYUD-v2 and PASCAL-context datasets shows that our method is superior to the state-of-the-art transformer-based multi-task learning techniques with higher accuracy and reduced computational resources. Moreover, our method is extended to video stream inputs, further reducing computational costs by efficiently sharing information across the temporal domain as well as the task domain. Our codes are available at \href{https://github.com/sarashoouri/EfficientMTL}{https://github.com/sarashoouri/EfficientMTL}.
\end{abstract}


\section{Introduction}
In various computer vision applications, it is required to obtain a comprehensive understanding of the visual scene by performing multiple tasks based on a single input image. These tasks often involve performing dense or pixel-wise predictions such as semantic segmentation, depth estimation, surface normal estimation, etc., for practical applications in autonomous driving, robotics, and augmented or virtual reality (AR/VR) \cite{ye2022inverted}. Traditionally, these tasks are tackled individually by training a separate neural network dedicated to each task. However, this single-task learning can lead to redundant computation and parameters, particularly for highly correlated tasks, losing the opportunity to perform faster inference as desired in real-time applications \cite{vandenhende2020mti}.
\figintro

Multi-task learning (MTL) \cite{caruana1997multitask,evgeniou2004regularized,li2022universal,zamir2020robust,zhang2021transfer} has been actively explored as a solution to this problem, which learns a single unified model to perform multiple tasks simultaneously. This approach allows the model to learn common representations and patterns from multiple supervised tasks \cite{zhang2021survey}, resulting in a more memory- and computation-efficient structure than employing multiple single-task networks. Additionally, multi-task networks can potentially enhance the performance of all tasks by leveraging the cross-task knowledge-sharing mechanism \cite{vandenhende2020mti}. Due to the benefits of multi-task learning, numerous multi-task structures have been proposed to improve task-wise interactions in dense visual scene understanding \cite{misra2016cross,xu2018pad,vandenhende2020mti, zhang2019pattern, zhou2020pattern, maninis2019attentive, gao2019nddr}. The recent success of transformer models in many downstream vision tasks \cite{liu2021swin, dosovitskiy2020image, zhu2020deformable, wang2020axial,ranftl2021vision,liu2022convnet} has led to the emergence of transformer-based multi-task networks \cite{seong2019video, bhattacharjee2022mult, ye2022inverted,chen2021pre,mohamed2021spatio}, aimed at improving multi-task performance even further.


However, MTL poses several potential drawbacks compared to single-task learning: (1) Simultaneously learning different tasks using a unified model can lead to unbalanced task competition and suboptimal performance for some tasks if the model fails to build a shared representation that generalizes to all tasks \cite{fifty2021efficiently}. (2) Balancing the loss between different tasks can be challenging due to the varying scales of task-specific loss terms, especially when the number of tasks increases \cite{chen2020just,chen2018gradnorm}. Although multiple task-balancing approaches have been proposed \cite{kendall2018multi,chen2018gradnorm,liu2019end} to address this problem, recent studies \cite{vandenhende2021multi} have shown that performance still becomes worse than individually-trained single-task networks. (3) MTL requires ground truth labels for all tasks per training sample, which is limiting because such annotations for certain tasks (e.g., semantic segmentation) may not always be available for the same input sample. It also often does not allow easily adding new tasks without retraining the entire model from scratch.

To alleviate the above limitations, we propose a solution that leverages the strength of both single-task and multi-task learning techniques, integrating the concept of knowledge-sharing. Our approach draws inspiration from \cite{chen2019toward}, which initially introduced the idea of efficient deep learning model communication through knowledge-sharing. In our work, we extend this concept to effectively execute multiple concurrent visual tasks taking the same input. We employ individually-trained single-task networks to maintain the desired performance and prevent unbalanced task competition. At the same time, we introduce a novel parameter- and computation-sharing strategy to facilitate knowledge-sharing and enhance inference efficiency. Our method focuses on transformer models that have shown outstanding results in vision tasks. First, we divide all tasks into a base task and multiple sub-tasks, where all task-specific networks adopt a common transformer structure as the \textit{backbone}. Next, we train a single network for the base task independently. Then, to share the inter-task information, we reuse \textit{not only weights but also activations} from the base task to train each sub-task. The weight-sharing concept is motivated by recent parameter-efficient transfer learning techniques \cite{houlsby2019parameter, guo2020parameter}. Specifically, we view the weights of sub-task models as the sum of weights from the base task and a \textit{delta weight} matrix, and apply $\ell_0$ regularization to encourage sparsity in the delta weight matrix as in the Diff-pruning \cite{guo2020parameter} approach. We then fix the positions of non-zero elements of the delta weight matrix and fine-tune them to make the \textit{activation difference} between the base task and sub-task also sparse by adopting $\ell_1$ regularization. As a result, the pre-computed activations from the base task can be shared and passed to sub-task networks during the inference. This reduces computation cost as the remaining operations for sub-tasks only involve sparse matrix-matrix multiplications, as shown in Figure \ref{fig:fig_intro}. Our proposed computation-sharing scheme significantly reduces the number of non-zero parameters for sub-tasks while allowing knowledge sharing between the main task and each sub-task. Extensive experiments on NYUD-v2 and Pascal-Context benchmarks demonstrate that our method outperforms state-of-the-art multi-task transformers, exhibiting fewer parameters and FLOPs counts to attain comparable or superior task accuracy.

Furthermore, we extend our method to the temporal domain to leverage the sparsity of differences between consecutive video frames, as shown in Figure \ref{fig:fig_intro}. Similar to the task domain, we employ $\ell_1$ regularization to force the activation differences between consecutive (time domain) frames to become sparse. As a result, during the inference of a sub-task at time $t$, it can exploit either the temporal domain or task domain sparsity to reuse activations from the same sub-task at time $t-1$ (temporal activation sharing) or the main task at time $t$ (task domain activation sharing). A simple strategy is then applied to combine these two sources of activation sharing for maximum efficiency and computation savings.

\begin{figure*}[t]
 \centering
 \includegraphics[width=2\columnwidth]{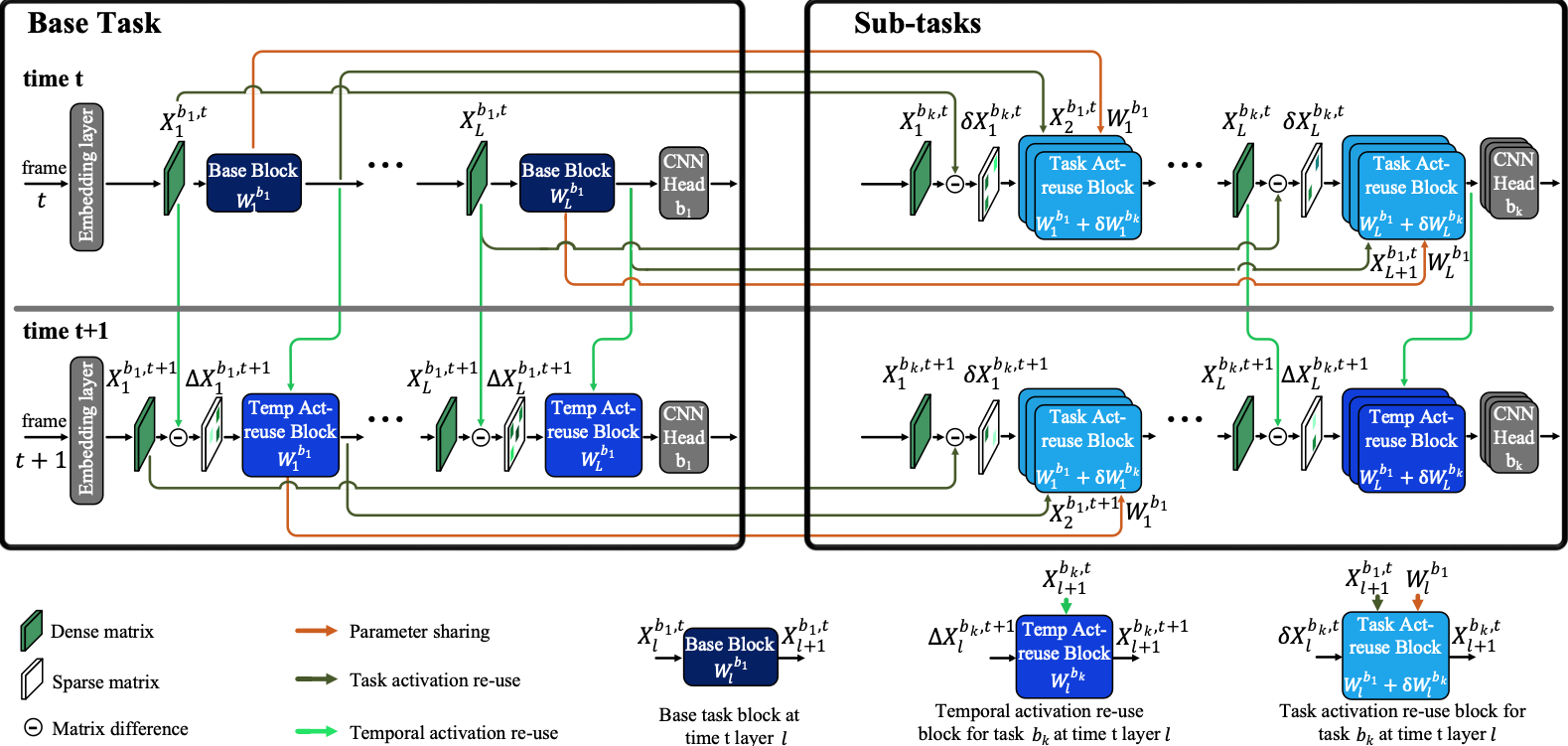}
 \centering
 \caption{Illustration of the proposed method for multi-task processing for video inputs. $X_{l}^{b_k,t}$, $\delta X_{l}^{b_k,t}$, and $\Delta X_{l}^{b_k,t}$ represent the input activation, delta task activation, and delta temporal activation of layer $l$ for task $b_k$ at time $t$, respectively.
 For a keyframe at $t$, the base task shares the parameter and activation computation with sub-tasks to improve the efficiency of the sub-task transformer models. For a non-keyframe at $t+1$, the base task first borrows the temporal activation from the previous frame at $t$ and then passes the calculated task activation to sub-tasks for layer $[1,l]$ at $t+1$. For the remaining layers of $[l+1,L]$ at $t+1$, the sub-tasks reuse the temporal activation from the previous time $t$ to further reduce the computation. }
 \label{fig:Structure}
\end{figure*}

Our contributions can be summarized as follows:
\begin{itemize}[noitemsep,topsep=0pt]

  \item We propose a novel activation-and parameter-sharing scheme to reduce the computation and storage redundancy to perform multiple dense/pixel-wise vision tasks with transformer models.

  \item We extend our computation-sharing method to video inputs and use a simple strategy to combine the activation-sharing sources between the task and temporal domains to maximize inference efficiency.

  \item We perform extensive experiments on NYUD-v2 and Pascal-Context datasets to quantify
the advantage of our proposed method in both performance and efficiency compared to prior multi-task learning methods. 
\end{itemize}

\section{Related Work}
\label{Related Work}
\subsection{Multi-task for Scene Understanding}
Multi-task learning (MTL) has made significant progress in several vision applications \cite{caruana1997multitask,evgeniou2004regularized,kumar2012learning,kendall2018multi}, such as joint object detection and semantic segmentation \cite{girshick2014rich, he2017mask}, and 3D recognition \cite{yang2016discriminative, yang2016latent}. Due to the benefits of MTL, various works have explored its use in multi-task dense scene understanding, where all the tasks require pixel-wise predictions \cite{eigen2015predicting, misra2016cross} based on convolutional neural networks (CNN) \cite{gao2019nddr, bruggemann2021exploring, maninis2019attentive}. Specifically, Cross-Stitch \cite{misra2016cross} enables soft feature fusion by utilizing a linear combination of the activations in each layer of task-specific networks. PAD-Net \cite{xu2018pad} first generates multi-modal auxiliary predictions and then uses a spatial attention strategy to distill information from the initial predictions. MTI-Net \cite{vandenhende2020mti} extends the idea of PAD-Net by proposing a multi-scale multi-modal distillation procedure to refine the feature propagation, leading to unique task interactions at each individual scale. On the other hand, PAP \cite{zhang2019pattern} and PSD \cite{zhou2020pattern} learn global and local affinities across tasks, and use them to refine the features for each task iteratively. Recently, the transformer model has also been applied to multi-task learning \cite{chen2021pre, mohamed2021spatio, seong2019video, bhattacharjee2022mult}, replacing CNNs for superior performance in several vision tasks \cite{liu2021swin,dosovitskiy2020image,bello2019attention,carion2020end,parmar2018image,ramachandran2019stand,touvron2021training,wang2018non}. For instance, InvPT \cite{ye2022inverted} leverages vision transformers as the backbone and explores self-attention for better spatial and task interaction, achieving state-of-the-art results for multi-task visual scene understanding.

Our approach differs from prior multi-task learning methods which mainly focus on enhancing cross-task interaction. Specifically, we utilize individually-trained single-task networks with a novel computation-sharing scheme to improve inference efficiency while preserving their performances to outperform prior multi-task learning methods.


\subsection{Parameter-Efficient Transfer learning}
Transfer learning and multi-task learning are two complementary machine learning paradigms that share a common goal of improving model efficiency through knowledge transfer. Transfer learning can be viewed as a specialized case of multi-task learning, where the knowledge (i.e., extracted features) is transferred from a source task or domain to improve performance on a related target task. This technique has been widely adopted in various applications, including computer vision \cite{caron2020unsupervised,noh2019transfer,sun2019meta}, natural language processing (NLP) \cite{gordon2020compressing,huang2021ghostbert,zhao2020masking,sanh2020movement,liu2019multi,stickland2019bert}, and speech recognition \cite{tomanek2021residual,thomas2022efficient}. The most common transfer learning methods for transformer-based models involve training a shared model, and then either fine-tuning model parameters or using linear probes with additional multi-layer perceptron (MLP) models \cite{wuunderstanding,tripuraneni2020theory,dufew} for individual tasks. The fine-tuning method can maintain high performance at the cost of high computational complexity for large transformers \cite{patterson2021carbon}, whereas linear probing exhibits relatively low performance with reduced complexity \cite{kumarfine,kornblith2019better,zhai2019large,he2020momentum}. Recently, there have been several research efforts focused on improving the parameter efficiency for transformers in NLP \cite{houlsby2019parameter, guo2020parameter, hu2021lora, zaken2021bitfit, karimi2021compacter}. For example, Adapter \cite{houlsby2019parameter} addresses this challenge by employing additional task-specific trainable modules after the feedforward network in each layer of the shared transformer architecture. Diff-pruning \cite{guo2020parameter} further improves parameter efficiency by learning sparse task-specific difference (`diff' or `delta') vectors to be added to the shared parameters. LoRA \cite{hu2021lora} takes a similar approach but decomposes the task-specific difference vector as a low-rank matrix. Parameter-efficient transfer learning has also been studied for vision transformers \cite{he2022parameter}. Unlike these prior methods that mainly focus on improving parameter efficiency, our approach extends the existing methods to improve both parameter and computation efficiency when multiple pixel-level tasks are performed on a single still image or consecutive video frames. 
\vspace{-2mm}

\section{Proposed Method}
The proposed approach is designed to enable the efficient processing of multiple dense (per pixel) tasks from a single input image. The approach is further optimized for video sequences to balance accuracy and efficiency, reducing computational costs by exploiting either the task or temporal domain sparsity as depicted in Figure \ref{fig:Structure}. We first formalize the problem for the algorithm and then present a general framework that leverages activation reuse across the task or temporal domain. 
\subsection{Framework Overview}
\label{framework}
\paragraph{Problem formulation: } The proposed multi-task model, $\mathcal{G}$, is designed to perform a set of $m$ visual tasks, $\mathcal{B}=\{b_1, b_2,\ldots, b_m\}$, on a single input image $I$. It generates the task outputs, $\hat{y}=\{\hat{y}^{b_1},\hat{y}^{b_2},\ldots,\hat{y}^{b_m}\}$, for each task in $\mathcal{B}$. Here we assume per-pixel tasks such as semantic segmentation or depth estimation. The model can be further applied to a set of video frames $\{I^{1},\ldots \, I^{T}\}$, where $I^{t}$ is the input frame at time $t$. Hence, the model output for the frame at time $t$ can be represented as $\mathcal{G}(I^{t}) \to \hat{y}^{t}=\{\hat{y}^{b_1,t},\hat{y}^{b_2,t},\ldots,\hat{y}^{b_m,t}\}$.

We begin by dividing the desired tasks into a base task and the other sub-tasks. Without loss of generality, suppose $b_1$ is the base task and the remaining tasks, $\{b_2,\ldots,b_m\}$, are the sub-tasks.
To perform each task, we define separate single-task networks and leverage transfer learning techniques to initialize the parameters of the sub-tasks using the base model and then fine-tune each model for a task-specific objective. 

Suppose a training dataset $\mathcal{D}^{b_k}=\{I_{n},y_{n}\}^{N}_{n=1}$ is given for a single network for task $b_k$. We seek to determine the optimal parameters $w^{b_k}$ for that model by solving the following optimization problem:
\setlength{\belowdisplayskip}{8pt}
\begin{equation}
\label{finetune_loss}
\setlength{\abovedisplayskip}{3pt} 
       \min_{w^{b_k}} \frac{1}{N} \sum_{n=1}^{N} \mathcal{C}^{b_k
}(f^{b_k}(I_{n};w^{b_k}),y_{n})+ \lambda \mathcal{R}(w^{b_k}),
\end{equation}
where $f^{b_k}(.;w^{b_k})$ represents a task-specific network, $\mathcal{C}^{b_k}(.)$ is a task-specific loss function, and $\mathcal{R}(.)$ is an optional regularization term with a hyperparameter $\lambda$. For the network $f^{b_k}$, we use a vision transformer \cite{dosovitskiy2020image,liu2021swin} as the backbone, $E^{b_k}$, followed by a small task-specific CNN head, $H^{b_k}$. That is, $f^{b_k}(.) = H^{b_k}(E^{b_k}(.))$ holds. While it is beneficial to have a separate model for each sub-task, such an approach can be inefficient due to the high memory and computation requirements during both training and inference, making it impractical for resource-constrained real-world applications. To address this issue, our method applies delta weight, delta temporal activation, and delta task activation pruning techniques to the $E^{b_k}$ for all existing sub-tasks as described in Sec. \ref{weigh_sharing_desc} and \ref{act_sharing_desc}. The CNN head $H^{b_k}$ for each sub-task is trained without exploiting delta weight/activation sparsity.
\vspace{-3mm}
\paragraph{Transformer backbone $E^{b_k}$: }
The transformer model consists of two fundamental computation elements in each layer: a multi-head self-attention module and a feedforward network (FFN). The multi-head self-attention module calculates the inter-dependencies between different positions in the input data through the following equations:
\vspace{-1mm}
\begin{gather}
\scalebox{0.91}{$
   \begin{split}
    MultiHead(Q,K,V)=Concat(head_1,\ldots,head_h)W_O\\
    \text{where} \: \: \: head_i=Attention(XW_Q, XW_K, XW_V), \: \: \: \: \: \: \: \:  \: \: \: \: \,\\
    Attention(Q,K,V)=softmax(QK^\top/ \sqrt D)V,  \: \: \: \: \: \: \: \: \: \: \: \: \quad \,
   \end{split}$}
\end{gather}
where $X\in R^{P\times D}$ is the activation input that has $P$ entries of $D$ dimension, $\{ W_Q, W_K, W_V\} \in R^{D\times D}$, and $W_O \in R^{Dh\times D}$ are model parameters, and $h$ is the number of heads. An FFN is applied to the multi-head output of the self-attention layers. It contains two linear projections $\{W_{F1}, W_{F2}\}\in R^{D\times D}$, connected by a GELU \cite{hendrycks2016gaussian} non-linearity $\sigma$, and it can be expressed as $FFN(X)=\sigma(XW_{F1})W_{F2}$. Therefore, each transformer block contains $4h + 2$ linear projections, each with a computational complexity of $O(PD^2)$. On the other hand, the calculation of the self-attention module, represented as $Attention(.)$, has a complexity of $O(P^2D)$. As such, reducing the computations for linear projections can be equally important as reducing the calculation of the self-attention module. The objective of our technique is to reduce the complexity of linear projections.
\raggedbottom
\vspace{-3mm}
\paragraph{Framework: } For simplicity, we describe the proposed activation reuse method for an arbitrary linear projection of a vision transformer block. Given the input activation $X^{b_1,t}_{in}$ and weight matrix $W^{b_1}$ for the base task $b_1$ at time $t$, the model first generates the outputs without activation reuse such that $X^{b_1,t}_{out}=X^{b_1,t}_{in}\, W^{b_1}$. For sub-task $b_k$, the model first learns a sparse task-specific delta weight matrix $\delta W^{b_k}$ and forms the final task-specific weight as $W^{b_k}=W^{b_1}+\delta W^{b_k}$, reusing the weights from the main task. This approach reduces the storage requirements of each sub-task by only storing the sparse $\delta W^{b_k}$ instead of the dense matrix $W^{b_k}$. However, the activation computation for $b_k$ involves $X^{b_k,t}_{out}=X^{b_k,t}_{in} \, (W^{b_1}+\delta W^{b_k})$, which has the same (or higher due to weight additions) complexity as the base task. To reduce the complexity, the computation is reformulated by introducing a delta task activation matrix, $\delta X^{b_k,t}_{in}=X^{b_k,t}_{in} -X^{b_1,t}_{in}$ with a goal to remove inter-task redundancies by making $\delta X^{b_k,t}_{in}$ sparse. This allows for activation sharing from the base task $b_1$ to sub-task $b_k$, resulting in a more efficient calculation, which can be written as:
\setlength{\abovedisplayskip}{8pt}
\setlength{\belowdisplayskip}{8pt}
\begin{gather}
\label{acti_reuse}
\scalebox{0.89}{$
   \begin{split}
    X^{b_k,t}_{out}=(X^{b_1,t}_{in} +\delta X^{b_k,t}_{in})\, (W^{b_1}+\delta W^{b_k})\qquad \qquad \qquad \quad \:\\
       =X^{b_1,t}_{in}\, W^{b_1} + \delta X^{b_k,t}_{in} \, (W^{b_1}+\delta W^{b_k})
          +X^{b_1,t}_{in} \, \delta W^{b_k}. 
   \end{split}$}
\end{gather}

As expressed in Eq. (\ref{acti_reuse}), our method borrows the activation computation from the base task $X^{b_1,t}_{in}\, W^{b_1}$ for sub-tasks and executes only sparse matrix-matrix multiplications in Eq. (\ref{acti_reuse}) rather than performing a dense matrix-matrix multiplication $X^{b_k,t}_{in}\, W^{b_k}$. Thus, the new calculation of a sub-task is reduced to $X^{b_k,t}_{Re}=\delta X^{b_k,t}_{in} \, (W^{b_1}+\delta W^{b_k}) +X^{b_1,t}_{in} \, \delta W^{b_k}$, where $\delta W^{b_k}$ and $\delta X^{b_k,t}_{in}$ are sparse matrices. This not only increases the speed of executing the sub-tasks but also reduces the number of required (non-zero) parameters. Additionally, to ensure that the $X^{b_k,t}_{Re}$ remains sparse for the next layer, we set a per-block task-specific threshold $th^{b_k}$ such that the activations whose absolute values are less than the predefined threshold $th^{b_k}$ are set to zero before being passed to the next layer. The final output activation can be written as:
\vspace{-2mm}
\begin{align}
\label{acti_reuse_short}
\begin{split}
X^{b_k,t}_{out}=X^{b_1,t}_{out}+Q(X^{b_k,t}_{Re}, th^{b_k}), 
\end{split}
\end{align}
where $Q(X, th)$ is an operator to substitute all elements less than $th$ in $X$ with zero.

Eq. (\ref{acti_reuse}) and (\ref{acti_reuse_short}) enable the activation sharing across different task domains (from $b_1$ to $b_k$). The same activation-sharing idea can also be applied to the temporal domain so that activations are shared from time $t$ to time $t+1$ for the same task in $\mathcal{B}$. Given the input activation $X^{b_k,t}_{in}$ and weight matrix $W^{b_k}$ for an arbitrary task $b_k$, we first perform the linear projection at time $t$ to obtain $X^{b_k,t}_{out}=X^{b_k,t}_{in}\,W^{b_k}$. To reduce the complexity for the next time step $t+1$, we introduce a delta temporal activation matrix $\Delta X^{b_k,t+1}_{in}= X^{b_k,t+1}_{in} - X^{b_k,t}_{in}$ and exploit temporal redundancies by making $\Delta X^{b_k,t+1}_{in}$\footnote{We use $\Delta X$ and $\delta X$ for matrix differences across temporal and task domains, respectively.} sparse. Hence, the calculation of task $b_k$ at time $t+1$ can be optimized by borrowing activations from time $t$, as follows:

\begin{gather}
\label{acti_reuse_temp}
\scalebox{0.92}{$
   \begin{split}
X^{b_k,t+1}_{out}=X^{b_k,t+1}_{in} \, W^{b_k}=(X^{b_k,t}_{in} +\Delta X^{b_k,t+1}_{in})\, W^{b_k} \quad \, \, \:\\
       =X^{b_k,t}_{in}\, W^{b_k} + \Delta X^{b_k,t+1}_{in} \, W^{b_k}.\\
   \end{split}$}
\end{gather}
Similar to the task domain, we can remove the dense matrix-matrix multiplication $X^{b_k,t+1}_{in} \, W^{b_k}$ by reusing the activations from the previous time step $t$, and only process $\Delta X^{b_k,t+1}_{in} \, W^{b_k}$ which is a sparse matrix-matrix multiplication, as in Eq. (\ref{acti_reuse_temp}).

In the following sections, we will describe our strategy to prune the delta weight, delta task activation, and delta temporal activation to be as sparse as possible. Moreover, we will explain how the task- and temporal-domain activation sharings are combined to minimize the task complexity.
\subsection{Learning and pruning delta weight}
\label{weigh_sharing_desc}
To simplify, we consider the collection of all linear layers in a transformer block as a set, denoted by $\Phi=\{W_Q^{1}, W_K^{1}, W_V^{1},\ldots, W_Q^{h}, W_K^{h}, W_V^{h},W_O,W_{F1},W_{F2}\}$. Given the weights of the sub-task $b_k$ and the base task $b_1$, we aim to learn a sparse delta weight matrix that can reparameterize the task-specific model as $\Phi^{b_k}=\Phi^{b_1}+\delta \Phi^{b_k}$. In this reparameterization, the weight of the base task $\Phi^{b_1}$ remains fixed during the fine-tuning process, only updating $\delta \Phi^{b_k}$ for the target sub-task. To promote sparsity in $\delta \Phi^{b_k}$, we follow the approach of Diff-pruning \cite{guo2020parameter} and apply $\ell_0$ regularization to $\delta \Phi^{b_k}$. Hence, the optimization loss function in Eq. (\ref{finetune_loss}) is modified to:
\begin{gather}
\label{weight_loss}
\scalebox{0.94}{$
   \begin{split}
       \min_{\delta \Phi^{b_k}} \frac{1}{N} \sum_{n=1}^{N} \mathcal{C}^{b_k
}(f^{b_k}(I_{n};\Phi^{b_1}+\delta \Phi^{b_k}),y_{n})+ \mathcal{R}_w(\delta \Phi^{b_k}),
\end{split}$}
\end{gather}
where $\mathcal{R}_w(\delta \Phi^{b_k})$ is defined such that:
\setlength{\abovedisplayskip}{8pt}
\setlength{\belowdisplayskip}{8pt}
\begin{gather}
\scalebox{0.95}{$
   \begin{split}
       \mathcal{R}_w(\delta \Phi^{b_k})=\lambda_w \|vec(\delta \Phi^{b_k})\|_0 =\lambda_w \sum_{j} \mathbb{1}\{ \delta \Phi_{j}^{b_k}\neq 0\},
\end{split}$}
\setlength{\belowdisplayshortskip}{0pt}
\end{gather}
where $\delta \Phi_{j}^{b_k}$ is the $j^{th}$ element of $\delta \Phi^{b_k}$. Since this regularization term is non-differentiable, we adopt a gradient-based learning approach that uses a relaxed binary mask matrix as described in \cite{guo2020parameter,louizos2017learning}. This involves defining a binary mask $M^{b_k}$ for sub-task $b_k$ and relaxing it into a continuous space using a stretched Hard-Concrete distribution \cite{wang2019structured,maddisonconcrete}, allowing a differentiable gradient path. The resulting mask is then element-wise multiplied with the dense delta weight matrix $\delta \Phi^{b_k}$ to produce a sparsified version.
\tablesota
\tabelablationnpascal
\subsection{Pruning delta task and temporal activations}
\label{act_sharing_desc}
\textbf{Delta task and temporal activation:} Similarly, we define $\mathcal{X}$ as a set of the activation inputs of all linear layers in a transformer block. Hence, the activation input difference between the base task $b_1$ and the sub-task $b_k$ at time $t$ is represented by $\delta \mathcal{X}^{b_k,t}$. Our goal is to prune these activation differences to minimize inter-task redundancies, as discussed in Sec. \ref{framework}. To encourage the sparsity in $\delta \mathcal{X}^{b_k,t}$, we fix the position of non-zero elements of the delta weight $\delta \Phi^{b_k}$, and then fine-tune non-zero delta weights by applying regularization to the activation differences. As $\delta \mathcal{X}^{b_k,t}$ depends on both the input image and task-specific weight matrix, it is challenging to follow the same approach used in delta weight pruning and learn a fixed binary mask which applies to all inputs. Thus, as an alternative approach, we use $\ell_1$ regularization (also known as Lasso), which utilizes a Laplacian-like distribution to increase the amounts of small values. The delta task activation regularization $\mathcal{R}_{a1}$ with a coefficient of $\lambda_{a1}$ is defined as follows:
\vspace{0mm}
\begin{gather}
\scalebox{0.95}{$
   \begin{split}
       \mathcal{R}_{a1}(\delta \mathcal{X}^{b_k,t})=\lambda_{a1} \|vec(\delta \mathcal{X}^{b_k,t})\|_1 =\lambda_{a1} \sum_{j} |\delta \mathcal{X}^{b_k,t}_j |,
\end{split}$}
\end{gather}
where $\delta \mathcal{X}^{b_k,t}_j$ is $j^{th}$ element of $\delta \mathcal{X}^{b_k,t}$. 

The same technique extends to the temporal redundancy pruning (Sec. \ref{framework}) to sparsify the delta temporal activation within the same (sub-)task. Consider $\Delta \mathcal{X}^{b_k,t+\tau}$ as the activation input difference between time $t$ and $t+\tau$ for task $b_k$. To encourage sparsity in $\Delta \mathcal{X}^{b_k,t+\tau}$, we define the delta temporal activation regularization $\mathcal{R}_{a2}$ with a coefficient of $\lambda_{a2}$ such that $\mathcal{R}_{a2}(\Delta \mathcal{X}^{b_k,t+\tau})=\lambda_{a2} \|vec(\Delta \mathcal{X}^{b_k,t+\tau})\|_1$. We apply temporal regularization to $\tau \in  [-2,2]$ and reformulate the optimization problem accordingly:
\begin{gather}
\scalebox{0.95}{$
   \begin{split}
       \min_{\delta \Phi^{b_k}} \frac{1}{NT}\sum_{t=1}^{T} \sum_{n=1}^{N} \mathcal{C}^{b_k
}(f^{b_k}(I_{n}^t;\Phi^{b_1}+M^{b_k}\,\delta \Phi^{b_k}),y_{n}^t)\\
+\mathcal{R}_{a1}(\delta \mathcal{X}^{b_k,t})+\sum_{\tau=-2}^2 \mathcal{R}_{a2}(\Delta \mathcal{X}^{b_k,t+\tau}).
\end{split}$}
\end{gather}

\textbf{Activation re-use combination from both domains:}
\label{activation-combination}
To efficiently combine activations from both the task and temporal domains, we compare the computational complexities of two approaches per layer. While the base task can only reuse activations across the temporal domain, sub-tasks can access activations from both sources: using either $\delta X^{b_k,t}_{in}$ as in Eq. (\ref{acti_reuse}) or using $\Delta X^{b_k,t+1}_{in}$ as in Eq.(\ref{acti_reuse_temp}). Suppose the (average) density (ratio of non-zero values) of the delta weight, delta task activation, and delta temporal activation for layer $l$ is denoted by $S_w^l$, $S_{a1}^l$, and $S_{a2}^l$, respectively. Then, Eq. (\ref{acti_reuse}) requires $\small{(S_w^l+S_{a1}^l)PD^2}$ multiplications, while Eq. (\ref{acti_reuse_temp}) requires $\small{(S_{a2}^l)PD^2}$ multiplications. Thus, if $S_w^l+S_{a1}^l < S_{a2}^l$, it is more efficient to reuse activations across the task domain. Conversely, if $S_{a2}^l \leq S_w^l+S_{a1}^l$, it is more reasonable to reuse activations from the previous frame of the same (sub-)task. We observed that the first few layers tend to be more sparse in the task domain, while the remaining layers are more sparse in the temporal domain. After evaluating $S_w^l$, $S_{a1}^l$, and $S_{a2}^l$ for all layers in each sub-task $b_k$, we determine the layer boundary $l^{b_k}$ so that the task domain activation reuse is performed for layers $l \leq l^{b_k}$, and temporal domain activation reuse is utilized for the remaining layers.
\section{Experiments}
\textbf{Dataset:} We evaluate our algorithm on two popular scene understanding datasets, \textbf{NYUD-v2} \cite{silberman2012indoor} and \textbf{PASCAL-Context} \cite{chen2014detect}. NYUD-v2 contains 1,449 indoor scene images with annotations for semantic segmentation (40 classes), depth estimation, surface normal estimation, and edge detection tasks, including 795 images for training and 654 for testing. PASCAL-Context covers indoor and outdoor scenes and comprises 4,998 training and 5,105 testing images, providing the labels for human parsing, semantic segmentation (21 classes), saliency estimation, edge detection, and surface normal estimation. For video input, we limit our analysis to the NYUD-v2 dataset, as PASCAL-Context does not provide images for different time frames. 

\textbf{Evaluation Metrics:} As in InvPT \cite{ye2022inverted}, semantic segmentation and human parsing are evaluated with mean Intersection over Union (mIoU), surface normal estimation with mean error (mErr), depth estimation with root mean square error (RMSE) in millimeter, edge detection with optimal-dataset-scale F-measure (odsF), and saliency detection with maximum $F_1$ score (maxF). 

\textbf{Training Details:} We perform our experiments using the ViT-B transformer \cite{dosovitskiy2020image} pre-trained on ImageNet-22K \cite{deng2009imagenet} as the backbone, with a patch size of $16\times16$ pixels. Training on the NYUD-v2 dataset is conducted using a batch size of $64$, distributed across 5 NVIDIA A40 single-precision GPUs, taking approximately 24 hours. The AdamW optimizer \cite{loshchilov2017decoupled} is utilized, with a learning rate of $1\times 10^{-4}$ and a weight decay rate of $1\times 10^{-6}$. For the PASCAL-Context dataset, a batch size of $6$, a learning rate of $5\times 10^{-5}$, and a weight decay rate of $1\times 10^{-6}$ are used. {Training on this dataset is performed on 6 NVIDIA A40 single-precision GPUs, taking around 40 hours. Both datasets are trained using a polynomial learning rate scheduler \cite{zeiler2012adadelta}.
\figvis
\figwholepasfinal
\raggedbottom

Our approach considers `semantic segmentation' as the base task and all other tasks as sub-tasks. To train these tasks, we first train the base task and store the weights and intermediate activations as the base weights and activations. Then, we follow a three-step training procedure for the sub-tasks. First, the sparse delta weight matrices are learned for each sub-task with $\ell_0$ regularization on delta weights. Then, the model is fine-tuned for a few epochs, and the learned delta weight is updated to improve performance. Finally, the $\ell_1$ regularization is applied to the difference of intermediate activations for each batch, and the non-zero delta weights are updated to balance the performance and delta activation sparsity. For video inputs, the $\ell_1$ regularization is applied to both delta task and temporal activations.
\subsection{Single Image Evaluation For Multi-task}

\textbf{Model Baselines and Variants:} We define the following baseline and model variants for the evaluation: (i) \textbf{`Multi-task learning (MTL)'} represents a state-of-the-art (SOTA) baseline multi-task approach that comprises a shared encoder and multiple task-specific decoders which are jointly optimized. The current SOTA MTL baseline, InvPT \cite{ye2022inverted}, uses a transformer as the encoder, while others typically employ a CNN as the backbone. (ii) \textbf{`Single-task learning (ST)'} has a common transformer backbone plus a task-specific small CNN head model structure for each task which is independently trained for task-specific parameters (for both backbone and head models) without using the delta weight and delta activation pruning methods. (iii) \textbf{`Ours + weight sharing'} is similar to the ST model but adds delta weight pruning to share transformer weights between the base and sub-tasks. (iv) \textbf{`Ours + weight + act sharing'} indicates our proposed method that also utilizes transformer activation sharing in the task domain.

\textbf{Qualitative results:} Figure \ref{fig:fig_pas_vis} presents sample visualizations from the proposed model for the PASCAL-Context dataset. We compare these results with the current SOTA InvPT \cite{ye2022inverted} to demonstrate the superiority of our model. The visual comparison reveals that our model produces more accurate predictions compared to InvPT, particularly for semantic segmentation, human parsing, and edge detection tasks. Additional visualization results can be found in the supplemental material.

\tabletmpSOTAwholeNEW

\textbf{Quantitative results:} Evaluation results for PASCAL-Context dataset and NYUD-v2 dataset are summarized in Table \ref{PAS_SOTA} ($\emph{left}$) and Table \ref{PAS_SOTA} ($\emph{right}$), respectively. The tables illustrate that our method outperforms prior approaches, including InvPT \cite{ye2022inverted}. We compare the FLOPs, number of parameters, required memory to store model parameters, and performance of each task against the `ST' model and the SOTA transformer-based MTL method InvPT in Table \ref{table:share_pasca} (a) and (b). Notice that our base task (semantic segmentation) performance is similar to that of the `ST' method since it does not use weight or activation sharing. For the PASCAL-Context dataset, our proposed model (`weight + act sharing') outperforms InvPT while reducing the FLOPs and parameters by \bm{$49.44\%$} and \bm{$74.0\%$}, respectively, compared to the `ST' model. Table \ref{table:share_pasca} (a) specifies the FLOPs required for each base and sub-task, indicating that adding a new sub-task requires only \bm{$37.6\%$} FLOPs of the `ST' model. For the NYUD-v2 dataset, we first evaluate the model with a single image without enabling temporal activation reuse to isolate the gain of the task activation reuse strategy. Table \ref{table:share_pasca} (b) shows that our method reduces the FLOPs and parameters by \bm{$40.5\%$} and \bm{$66.6\%$} compared to the `ST' model while achieving comparable/better results than the InvPT baseline. To evaluate the required memory, we adopt the CSR (Compressed Sparse Row) method to store the sparse delta weight matrices. This approach efficiently stores only the non-zero elements and their corresponding positions, making it a suitable choice for our approach. The memory storage comparison results are presented in Table \ref{table:share_pasca}, further showcasing the memory-saving benefits of our approach.
Figure \ref{fig:all_sparsity} shows the overall delta weight and delta task activation sparsity for all sub-tasks in the Pascal-Context dataset ($\emph{left}$) and the NYUD-v2 dataset ($\emph{right}$). 
It reveals that human parsing and edge detection are more closely related to the base task, hence their sparsities are higher than other sub-tasks.

\textbf{Ablation study:} We explore the impact of $\ell_0$ and $\ell_1$ regularization coefficients on the performance, computation, and memory storage reduction for each task in both PASCAL-Context and NYUD-v2 datasets. To determine the optimal values of $\lambda_w$ and $\lambda_{a1}$, we conduct a comprehensive hyperparameter search, sweeping $\lambda_w$ in the range of $[1\times 10^{-8},10\times 10^{-8}]$ and $\lambda_{a1}$ in the range of $[1\times 10^{-10},10\times 10^{-10}]$. Figure \ref{fig:fig_abl_task} shows the performance of the human parsing task with different values of $\lambda_w$ and $\lambda_{a1}$. As expected, increasing $\lambda_w$ saves more parameters but reduces accuracy by $0.30\%$. Similarly, increasing $\lambda_{a1}$ saves more computations at the cost of decreased accuracy by $1.24\%$. For each sub-task, we select the optimal $\lambda_w$ and $\lambda_{a1}$ to balance computation/memory storage and task-specific performance.

In addition, we evaluate the impact of using a different backbone network by replacing the ViT-B transformer model with the Swin-B \cite{liu2021swin} model. This evaluation is performed for both the InvPT baseline and our proposed method with semantic segmentation (`Semseg') as the base task. Furthermore, we compare the performance of our method when human parsing (`Parsing') is selected as the base task, aiming to demonstrate the effectiveness of our approach when choosing an alternative task as the base. The results on the Pascal-context dataset are shown in Table \ref{SWINB}. When all approaches use Swin-B, our method (weight \& act sharing) with the `Semseg' as the base task outperforms InvPT baseline, exhibiting significantly reduced FLOPs and parameters (by \bm{$45.73\%$} and \bm{$71.07\%$}) compared to the `ST' model. Similarly, our method with the `Parsing' as the base task surpasses the InvPT baseline, exhibiting substantial reductions in FLOPs and parameters (by \bm{$42.16\%$} and \bm{$71.03.\%$}, respectively) compared to the `ST' model. It is also evident that the `Semseg' task delivers slightly better performance as the base task. We attribute this improvement to the fact that segmentation is a highly informative task, generating powerful activations and features that can be effectively generalized to other computer vision tasks.


\subsection{Video Frame Evaluation}
Now, we evaluate the temporal activation-sharing method using the NYUD-v2 dataset. As discussed in Sec \ref{activation-combination}, we first evaluate the sparsity ratios of all layers for a task and then determine the optimal layer boundary for sharing mode switching. Specifically, for task $b_k$, we leverage the task-domain activation sharing for the first $l^{b_k}$ layers and subsequently switch to the temporal-domain activation sharing for the remaining layers. Figure \ref{fig:task-temp-sparse} shows the sparsity comparison between the delta task and temporal activations for depth estimation, surface normal estimation, and edge detection tasks to determine the sharing mode switching boundary $l^{b_k}$.


As the frames in the NYUD-v2 dataset are sparsely annotated (e.g., only every 20th frame is annotated), we evaluate the performance of annotated ground truth (GT) frames by considering all possible interval offsets between the start (keyframe) and GT frames within the range of $[0,4]$ and record the averaged performance and FLOPs. To assess the impact of task and temporal activation combinations, we evaluate the following model variants: (i) \textbf{`ST'}, (ii) \textbf{`InvPT'}, (iii) \textbf{`+ Task act'} using only task domain activation reuse for all frames; and (iv) \textbf{`+ Task act + Temporal act'} which is the proposed model using the combination of temporal and task activation sharing strategy. The proposed model employs the same computation as the `+ Task act' model for a keyframe (it appears up to 4 frames earlier than the GT frame) but reduces the computation of base and sub-tasks for non-keyframes. Table \ref{tempPAS_SOTA} compares the performance and FLOPs of these variants. We observe that the proposed approach significantly reduces FLOPs by \bm{$42.3\%$} and \bm{$65.7\%$} compared to `+ Task act' and `ST' models, respectively. 
\begin{figure}[!h]
 \centering
       \includegraphics[width=1\columnwidth]{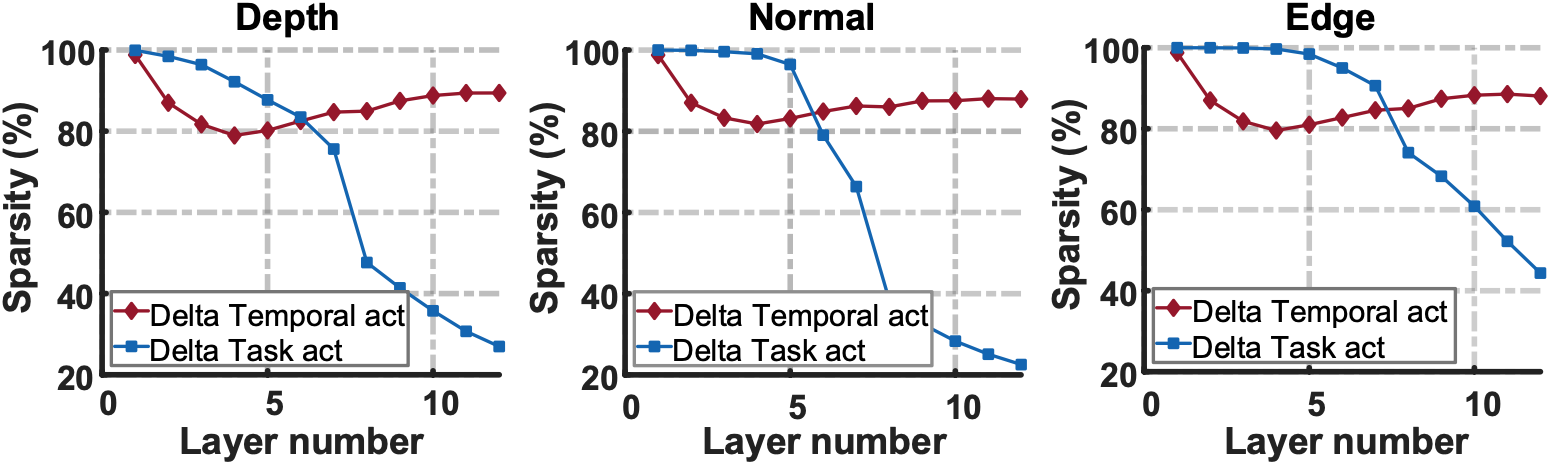}
         \caption{The sparsity comparison between delta task and temporal activation for NYUD-v2 dataset.}
    \label{fig:task-temp-sparse}
\end{figure}


\section{Discussion}
Calculating the actual inference speed of deep learning models on available computing platforms is a critical factor in evaluating performance. However, due to the utilization of sparse matrix-matrix multiplications in our proposed approach, measuring the true speedup requires a specialized sparse linear algebra processor or hardware accelerator, which is not yet readily available for commercial platforms. As a practical alternative, we analyze the required FLOPs to estimate the speedup potential. While FLOPs provide a reasonable comparison metric, we acknowledge that they may not fully represent the actual speed gains achieved. This calls for future work to develop a sparsity-aware transformer accelerator that will allow us to quantify the true speedup achieved by our approach.

\section{Conclusion}
This paper presents a novel computation- and parameter-sharing scheme for transformer-based multiple visual tasks that are concurrently performed on the same input. Motivated by recent transfer learning techniques, our scheme reuses the weights and activations of the base task by training sub-tasks with sparse weight and activation differences via $\ell_0$ and $\ell_1$ regularization. As a result, the activations from the base task can be shared with all sub-tasks, reducing both parameter and computation redundancy significantly. Additionally, the proposed scheme is extended to video inputs to further reduce computation redundancy in the temporal domain. Evaluation results confirm that our method attains better/comparable performance with fewer parameters and FLOPs than state-of-the-art multi-task learning methods. 
\vspace{-2mm}

\vspace{-2mm}
\paragraph{Acknowledgements.} This work was supported in part by COGNISENSE, one of seven centers in JUMP 2.0, a Semiconductor Research Corporation (SRC) program sponsored by DARPA. The authors would like to thank Morteza Tavakoli Taba, Bowen Liu, and Changwoo Lee at the University of Michigan for their valuable discussions.

{\small
\bibliographystyle{ieee_fullname}
\bibliography{egbib}
}

\end{document}